\documentclass{isprs} 
\usepackage{setspace}
\usepackage{geometry} 
\usepackage{epstopdf}
\usepackage[labelsep=period]{caption}  
\usepackage[british]{babel} 
\usepackage[hang]{footmisc}
\usepackage{natbib}
\usepackage{subcaption}
\usepackage{siunitx}
\usepackage{graphicx}
\usepackage{arydshln} 
\usepackage{booktabs}
\usepackage{xcolor}
\usepackage{multirow}
\usepackage{adjustbox}
\usepackage{enumitem}
\usepackage{svg}
\usepackage[hang]{footmisc}
\usepackage{url}

\newcommand{\Point}[1]{{\ensuremath{\textbf{\textit{#1}}}}}
\newcommand{\Vector}[1]{{\ensuremath{\textbf{\textit{#1}}}}}
\newcommand{\Matrix}[1]{{\ensuremath{\textnormal{\textbf{\textit{#1}}}}}}
\newcommand{\TMatrix}[3]{{{}^{\mathrm{#1}}\Matrix{#2}_{\mathrm{#3}}}}

\begin{document}

\title{Novel View Synthesis with Neural Radiance Fields \\ for Industrial Robot Applications}
\date{}


\author{
    \begin{tabular}{c}
        Markus Hillemann\textsuperscript{1}, Robert Langend\"orfer\textsuperscript{1}, Max Heiken\textsuperscript{2}, Max Mehltretter\textsuperscript{2},  Andreas Schenk\textsuperscript{1}, \\
        Martin Weinmann\textsuperscript{1}, Stefan Hinz\textsuperscript{1}, Christian Heipke\textsuperscript{2}, Markus Ulrich\textsuperscript{1}
    \end{tabular}
 }

\address{\textsuperscript{1}Institute of Photogrammetry and Remote Sensing, Karlsruhe Institute of Technology, Germany - \\(markus.hillemann@, robert.langendoerfer@student., andreas.schenk@, \\ martin.weinmann@, stefan.hinz@, markus.ulrich@)kit.edu\\
    \textsuperscript{2}Institute of Photogrammetry and GeoInformation, Leibniz Universit\"at Hannover, Germany -\\ (heiken, mehltretter, heipke)@ipi.uni-hannover.de}



\abstract{Neural Radiance Fields (NeRFs) have become a rapidly growing research field with the potential to revolutionize typical photogrammetric workflows, such as those used for 3D scene reconstruction. As input, NeRFs require multi-view images with corresponding camera poses as well as the interior orientation. In the typical NeRF workflow, the camera poses and the interior orientation are estimated in advance with Structure from Motion (SfM). But the quality of the resulting novel views, which depends on different parameters such as the number and distribution of available images, the accuracy of the related camera poses and interior orientation, but also the reflection characteristics of the depicted scene, is difficult to predict. In addition, SfM is a time-consuming pre-processing step, and its robustness and quality strongly depend on the image content. Furthermore, the undefined scaling factor of SfM hinders subsequent steps in which metric information is required.
In this paper, we evaluate the potential of NeRFs for industrial robot applications. To start with, we propose an alternative to SfM pre-processing: we capture the input images with a calibrated camera that is attached to the end effector of an industrial robot and determine accurate camera poses with metric scale based on the robot kinematics. We then investigate the quality of the novel views by comparing them to ground truth, and by computing an internal quality measure based on ensemble methods. For evaluation purposes, we acquire multiple datasets that pose challenges for reconstruction typical of industrial applications, like reflective objects, poor texture, and fine structures. We show that the robot-based pose determination reaches similar accuracy as SfM in non-demanding cases, while having clear advantages in more challenging scenarios. We also report results of the novel view quality for different NeRF approaches, showing that an additional online pose refinement may be disadvantageous. Finally, we present first results of applying the ensemble method to estimate the quality of the synthetic novel view in the absence of a ground truth.
}


\maketitle


\section{Introduction}\label{sec:Introduction}
Neural Radiance Fields (NeRFs) \citep{mildenhall2020nerf} have become a rapidly growing research field. Given a set of oriented images, NeRFs allow the generation of novel views of a learned 3D scene with impressive fidelity to reality. 
In industrial applications, such novel views could be used, for instance, to efficiently generate photorealistic training images for deep-learning-based approaches like 6D object pose estimation in bin-picking applications or image-based detection of anomalies. In general, the quality of novel views depends on the accuracy of the input data \citep{lin2021barf, jager2023density}. In the typical workflow, the camera poses and the interior orientation of the camera are estimated in advance with established Structure from Motion (SfM) approaches, such as COLMAP \citep{schonberger2016structure}. However, SfM is sensitive to the quality of local registration, easily converges to suboptimal solutions \citep{lin2021barf}, and sufficient and well-dis\-tributed texture in the images is a prerequisite for accurate results.

In this paper, we evaluate the potential of novel view synthesis with NeRFs for industrial robot applications: An industrial robot is used to acquire the input images with a camera that is mounted at its end effector. For each acquired image that is used as input for NeRF training, the robot pose is queried from the robot controller. In order to calculate the respective camera poses, the pose between the tool coordinate system and the camera coordinate system (hand--eye pose) must be known. It is determined by performing a hand--eye calibration. The interior orientation of the camera is calibrated simultaneously with the hand--eye calibration. Consequently, due to the very high accuracy of modern industrial robots and target-based calibration methods, the accuracy of our input data for NeRFs is high.

In addition, we investigate the quality of the novel views by comparing them to ground truth using the established metrics Peak Signal to Noise Ratio (PSNR)\footnote{\url{https://lightning.ai/docs/torchmetrics/stable/image/peak_signal_noise_ratio.html} (last access 29/04/2024)} and Structural SIMilarity (SSIM) \citep{wang2004SSIM} incl.\@ their respective standard deviations across the evaluation images for different industrial scenes and different backgrounds.
Because many industrial applications benefit from knowing the uncertainty of the results, we further compute an internal quality measure of the novel views based on ensemble methods in order to be able to predict the quality of novel views without a ground truth image. 


\section{Related work}\label{sec:RW}
The original concept of NeRF \citep{mildenhall2020nerf} is to represent a scene as continuous radiance field, where density and color are a function of position and viewing angle. This function is commonly parameterized in terms of a neural network. 
Images of a scene can be estimated from the radiance field via volumetric rendering: Each pixel of an oriented image is regarded as a ray intersecting the scene, where the color of a pixel is estimated by integrating over the density and color along the ray. During training, this concept is applied to render images for which there is a real counterpart in the training data. The photometric error, i.e., the per pixel color error between the rendered and the real image, is used to optimize the network parameters. \citet{barron2021mip} improve upon this method by regarding pixels as cones instead of rays, which has an anti-aliasing effect and increases rendering quality. This improvement is also adopted by Nerfacto \citep{tancik2023nerfstudio}, which additionally incorporates further extensions from various other NeRF methods, such as the refinement of the camera poses from Bundle-Adjusting Neural Radiance Fields (BARFs) \citep{lin2021barf}, with the overall goal of creating a compromise between quality and training time.


Inspired by NeRFs, \citet{kerbl20233d} developed 3D Gaussian Splatting (3DGS), which is currently considered state-of-the-art in novel view synthesis in terms of computation time and quality. 3DGS also uses volumetric rendering to generate synthetic counterparts to real training images and uses the photometric error as loss to optimize the scene representation. However, instead of representing the scene as a continuous radiance field encoded in the weights of a neural network, 3DGS uses a set of continuous 3D Gaussians. 
The continuous representation with 3D Gaussians allows for optimization due to its differentiability. At the same time, the scene is stored efficiently by representing the 3D Gaussians as discrete ellipsoids with opacity and view-dependent color information. Typically, as in other NeRF approaches, SfM is employed to determine the required camera poses and interior orientation. The sparse SfM point cloud is then used to initialize the set of 3D Gaussians, which is subsequently densified and improved upon in an iterative optimization.
Like NeRFs, 3DGS can therefore be considered as an alternative to Multi-View Stereo (MVS) approaches for dense 3D scene reconstruction in the typical photogrammetric pipeline of SfM followed by MVS, see e.g. \citep{Hackstein2024} for a comparison of the resulting 3D surface and dense matching results. However, unlike MVS, the aim of 3DGS is not to generate a 3D scene reconstruction that is as dense and accurate as possible, but to find a compact representation of the 3D scene in the form of 3D Gaussians that allows novel views to be rendered as realistically as possible. 

\citet{lin2021barf} show that the accuracy of the camera poses limits the quality of a radiance field and thus of rendered images.
As the camera poses obtained via SfM are often not accurate enough, especially in the presence of poor or repetitive texture, some NeRF-based approaches \citep{lin2021barf, jeong2021self, wang2021nerf, bian2023nope, tancik2023nerfstudio} jointly optimize the radiance field and the camera poses. 
However, these approaches still require initial values for the camera poses, or employ a coarse-to-fine schedule to try to avoid getting stuck in local minima, while \citet{bian2023nope} depend on matching point clouds of monocular depth maps. Either strategy may fail in the industrial context.
Furthermore, the simultaneous estimation of camera poses and the radiance field introduces an additional layer of ambiguity, leading to a slower convergence and a less stable optimization \citep{bian2023nope}.
In the present work, we address this problem by computing the camera poses directly from the known robot poses and the calibrated hand--eye pose. 

The DTU dataset \citep{aanaes2016large} provides images that are often utilized to characterize the performance of NeRFs. The images were captured with an industrial robot arm and accurate camera poses are provided. A set of camera poses was determined in advance with a high accuracy using a calibration object. For image acquisition, the robot was moved to the same set of poses again. As a result, the accuracy of the camera poses depends on the accuracy of the determined poses of the calibration object and the pose repeatability of the robot, which typically is very high. While this approach provides a high-quality dataset, its use in practical robot applications is limited, because it is impossible to flexibly choose the camera poses. Consequently, distribution and number of camera poses cannot be automatically chosen, for example based on the object's characteristics, nor adapted, for example based on previously occluded object parts. In contrast, our approach does not rely on pre-calibrated camera poses but allows to use poses that are optimal for a specific application. Since the absolute pose accuracy of industrial robots is typically lower than its pose repeatability \citep{placzek:18}, our approach may lead to less accurate poses overall compared to the DTU dataset. However, we consider the proposed data acquisition scheme to be more realistic and flexible for industrial applications.

\citet{jager2023density} study the impact of pose accuracy on the quality of rendered images by simulating noisy poses. 
\citet{vslapak2023neural} also study the applicability of NeRFs in the industrial domain, but rely exclusively on synthetic data. However, simulations like these are simplifications of reality and the transferability to real scenarios remains open.




Lastly, most NeRF-based methods model a deterministic function that does not allow estimating the uncertainty associated with the generated synthetic images.
\citet{shen2022conditional} address this limitation and approximate the posterior distribution of the radiance field based on the training images. Propagating this information through the volumetric rendering, the uncertainty of the rendered color can be estimated per pixel in a synthetically generated image.
\citet{sunderhauf2023} instead employ ensemble-based uncertainty estimation, which is more straight-forward to implement and can be applied to various NeRF architectures.
Knowledge of the uncertainty associated with an estimate is of particular importance in industry.
However, to the best of the author's knowledge, the potential of uncertainty-aware NeRF-based methods has not yet been investigated in the context of industrial applications.
Thus, we particularly focus on this aspect in our NeRF uncertainty investigations.


\section{Setup of the industrial robot application}\label{sec:setup}
\begin{figure}[t]
  \centering
  \begin{subfigure}[t]{0.39\columnwidth}
    \includegraphics[width=\textwidth]{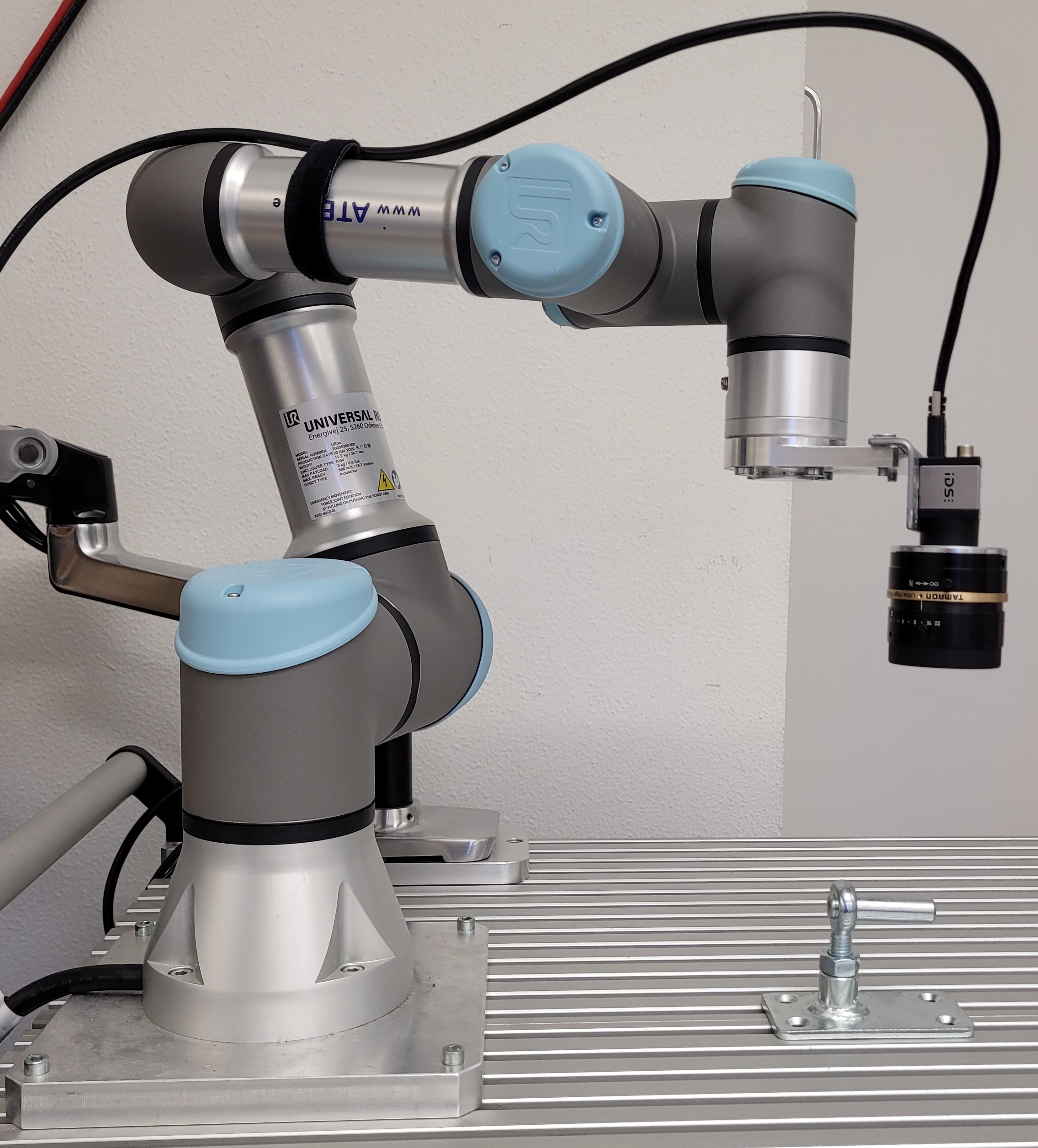}
    \caption{}
  \end{subfigure}
  \begin{subfigure}[t]{0.6\columnwidth}
    \includesvg[width=\textwidth]{figures/coordinate_systems_tex.svg}
    \caption{}
  \end{subfigure}
  \caption{(a) Hardware setup and data capturing. (b) Relevant 3D coordinate systems, observed poses (green), and poses calculated during hand--eye calibration (red); adapted from \citet{ulrich2024uncertainty}.}
  \label{fig:setupAndCOS}
\end{figure}

Our industrial robot setup consists of an IDS U3-31J0CP-C-HQ RGB camera with a 8\,mm Tamron M23Fm08 lens rigidly attached to the end effector of a Universal Robots UR3e robot arm, as shown in Figure~\ref{fig:setupAndCOS}(a). The UR3e has a maximum reach of 500\,mm, is calibrated by the manufacturer, and has a pose repeatability per ISO 9283 of $\pm0.03\,$mm. The RGB camera has a resolution of $2855\times2848\,$px and the lens has a field of view of $58.5$\textdegree $\times 45.1$\textdegree$\,$  as well as a minimum object distance of $0.1\,$m. 

We use the HALCON-implementation\footnote{\url{www.mvtec.com/products/halcon} (last access 29/04/2024)} of the uncertainty-aware hand--eye calibration approach of \cite{ulrich2024uncertainty} to simultaneously calculate the interior orientation of the camera and the hand--eye pose. The approach minimizes the reprojection error of points on a calibration object that was acquired by the camera in multiple different robot poses. 
The uncertainty of the robot, which considers pose repeatability and absolute pose accuracy, is explicitly taken into account in this approach in order to increase the accuracy of the estimated hand--eye pose. We capture $N=32$ images in total for calibration.

The result of hand--eye calibration is reported in terms of the root mean square translation (RMST), root mean square rotation (RMSR), and reprojection root mean square (RRMS) differences. Let $\TMatrix{b}{H}{a}$ denote the 4$\times$4 homogeneous transformation matrix that represents a rigid 3D transformation of points from coordinate system a to b. The basis for calculating RMST, RMSR, and RRMS is the typical chain of transformations for hand--eye calibration \citep{strobl2006optimal}
\begin{equation} \label{eq:HE_loop}
    \TMatrix{b}{H}{w}=\TMatrix{b}{H}{t}_{,j} \TMatrix{t}{H}{c} \TMatrix{c}{H}{w}_{,j}\mathrm{,}
\end{equation} 
with the coordinate systems world ($\mathrm{w}$), camera ($\mathrm{c}$), tool ($\mathrm{t}$), and base ($\mathrm{b}$), the $j$-th robot pose $\TMatrix{b}{H}{t}_{,j}$, and camera pose $\TMatrix{c}{H}{w}_{,j}$, as well as the calibrated hand--eye pose $\TMatrix{t}{H}{c}$ and world--base pose $\TMatrix{b}{H}{w}$ as shown in Figure~\ref{fig:setupAndCOS}(b). 
With the discrepancy matrix 
\begin{equation}
    \Delta\Matrix{H}_j = (\TMatrix{c}{H}{t} \TMatrix{t}{H}{b}_{,j} \TMatrix{b}{H}{w})^{-1} \TMatrix{c}{H}{w}_{,j}
\end{equation}
in this transformation chain, RMST and RMSR are calculated as follows:
\begin{equation} \label{eq:RMST}
    \mathrm{RMST} = \sqrt{\frac{1}{N} \sum_{j=1}^{N} \mathrm{trans}(\Delta\Matrix{H}_j)^2}
    \enspace,
\end{equation}
\begin{equation} \label{eq:RMSR}
    \mathrm{RMSR} = \sqrt{\frac{1}{N} \sum_{j=1}^{N} \mathrm{rot}(\Delta\Matrix{H}_j)^2}
    \enspace.
\end{equation}
Here, trans($\cdot$) calculates the length of the translation component and rot($\cdot$) the angle of the Rodrigues rotation of a matrix.
The RRMS is calculated by the root mean square of the differences of the observed image points $\Point{p}_{j,k}$ and the 3D points $\Point{p}_k$ ($k=1,\ldots,M$) of the calibration object projected into the image plane:
\begin{equation} \label{eq:RRMS}
  \mathrm{RRMS} = \sqrt{\frac{1}{N} \sum_{j=1}^{N} \frac{1}{M} \sum_{k=1}^{M} \left(\Point{p}_{j,k} -  \pi (\TMatrix{c}{H}{w}_{,j} \cdot \Point{p}_{k}, \Vector{i})\right)^2}
  \enspace ,
\end{equation}
where $\pi (\Point{p}_\mathrm{c}, \Vector{i})$ is the projection of the point $\Point{p}_\mathrm{c}$, which is given in the camera coordinate system, into the image, $\Vector{i}$ is the interior orientation, and
\begin{equation} \label{eq:cameraHworld}
    \TMatrix{c}{H}{w}_{,j} = \TMatrix{c}{H}{t} \TMatrix{t}{H}{b}_{,j} \TMatrix{b}{H}{w}
\end{equation}
is the camera pose, which is calculated based on the transformation chain.

The values after hand--eye calibration are $\mathrm{RMST}=0.11$\,mm, $\mathrm{RMSR}=0.028$\textdegree, and $\mathrm{RRMS}=1.00\,$px. For comparison, we also report the RRMS of a standard camera calibration with the same data, which is $\mathrm{RRMS}=0.33\,$px. Here, the camera poses $\TMatrix{c}{H}{w}_{,j}$ are estimated directly and not based on the transformation chain via the robot.
Moreover, we report the mean standard deviations of the robot poses used for hand--eye calibration. They are $\sigma_t=0.07$\,mm for translation and $\sigma_R=0.017$\textdegree for rotation, which is in the same order of magnitude. Consequently, the accuracy of the robot poses corresponds to approximately two to three times the pose repeatability of the robot. This is in line with previous results \citep{ulrich2024uncertainty}. 


\section{Experimental results}\label{sec:Experiments}

In this section, details of the data capturing are given, and the accuracy of the camera poses calculated with our procedure as well as with the conventional procedure using SfM is assessed. Subsequently, details of the evaluated NeRFs are described before their results are presented quantitatively and qualitatively on the basis of a meaningful selection of examples. Finally, exemplary results of the uncertainty quantification are presented.

\subsection{Datasets}\label{sec:datasets}
\begin{figure}[t]
  \centering
  \begin{subfigure}[b]{0.31\columnwidth}
    \includegraphics[width=\textwidth,angle=180,origin=c]{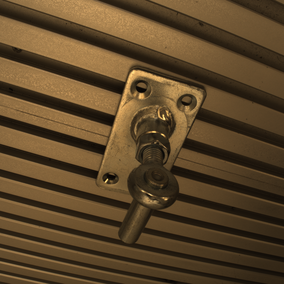}
    \caption{Dataset 1}
  \end{subfigure}
  \begin{subfigure}[b]{0.31\columnwidth}
    \includegraphics[width=\textwidth,angle=180,origin=c]{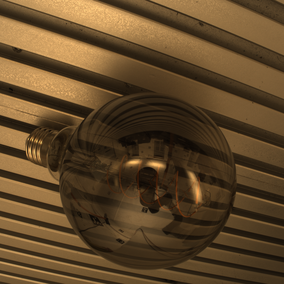}
    \caption{Dataset 2}
  \end{subfigure}
  \begin{subfigure}[b]{0.31\columnwidth}
    \includegraphics[width=\textwidth,angle=180,origin=c]{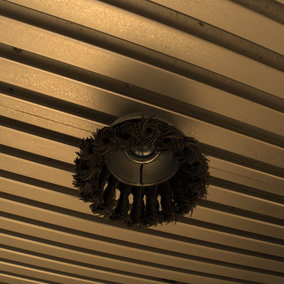}
    \caption{Dataset 3}
   \end{subfigure}

  \begin{subfigure}[b]{0.31\columnwidth}
    \includegraphics[width=\textwidth,angle=180,origin=c]{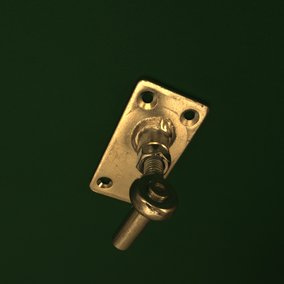}
    \caption{Dataset 4}
  \end{subfigure}
  \begin{subfigure}[b]{0.31\columnwidth}
    \includegraphics[width=\textwidth,angle=180,origin=c]{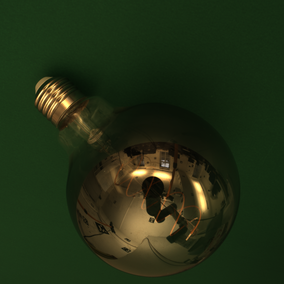}
    \caption{Dataset 5}
  \end{subfigure}
  \begin{subfigure}[b]{0.31\columnwidth}
    \includegraphics[width=\textwidth,angle=180,origin=c]{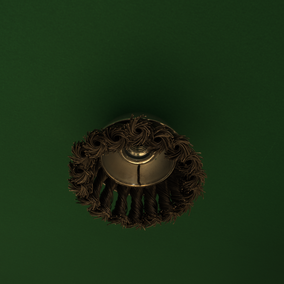}
    \caption{Dataset 6}
   \end{subfigure}

  \caption{Example image of each dataset.}
  \label{fig:objects}
\end{figure}

We capture images of three different industry-related objects (see Figure \ref{fig:objects}). In principle, our approach allows us to select the optimal poses for each object individually without additional computational effort. For the evaluations in this paper, however, we select the same systematically arranged poses for each dataset. The camera poses are sampled on a hemisphere with radius $R=0.2$\,m, with the camera always pointing approximately to the center of the object. The camera is upside down, as this allows a larger radius. For the sake of clarity, all shown images of the datasets are rotated by 180\textdegree. The sampling interval is $d_{\mathrm{lat}}=5^\circ$ in latitude and $d_{\mathrm{lon}}=5^\circ$ in longitude with an elevation between 55 and 85\textdegree, which results in 504 images per dataset. All objects have poor texture and a (partially) reflective surface. Additionally, the light bulb (Figures~\ref{fig:objects}(b) and (e)) is partly transparent and the metal brush (Figures~\ref{fig:objects}(c) and (f)) contains fine structures. These characteristics pose challenges for image processing in general and novel view synthesis with NeRFs in particular, and allow us to explore the limits of NeRFs. At the same time, the objects are relevant in practice, since they are typical for industrial applications. 

In total, we capture six different datasets, whereby each object is captured with two different backgrounds: An aluminum profile and a homogeneous green background. The aluminum profile provides a good fine-grained texture and thus allows for accurate SfM results. As already mentioned, such a good texture is often not available in industrial applications. Therefore, we also collect data with a homogeneous background on which no feature points can be extracted. In such cases, the estimation of the camera poses with SfM relies on the texture of the objects.

\subsection{Pose accuracy}\label{sec:pos_accuracy}
To analyze the accuracy of the camera poses determined with COLMAP and the camera poses approached with the robot, we compare them to a ground truth (GT). For this GT, the poses are determined in a third procedure with higher accuracy by using a calibration object and spatial resection. To demonstrate the higher accuracy of the GT, the RRMS is shown in Table \ref{tab:pose_acc} for each of the three procedures. The calculation of the RRMS is based on Equation (\ref{eq:RRMS}). Note that $\Point{p}_k$ are points of the calibration target for the GT and the robot, and points of the sparse point cloud for COLMAP. In addition, for the GT and COLMAP, the camera poses $\TMatrix{c}{H}{w}_{,j}$ are calculated directly, whereas for the robot it is calculated via the transformation chain with Equation (\ref{eq:cameraHworld}). The RRMS for the GT is considerably smaller compared to the COLMAP and the robot variant. Further, Table~\ref{tab:pose_acc} shows the Mean Translation Error (MTE) and the Mean Rotation Error (MRE) of COLMAP and the robot compared to GT. MTE and MRE\footnote{cf.\ translation\_part and rotation\_angle\_deg of APE at \url{https://github.com/MichaelGrupp/evo/blob/master/notebooks/metrics.py_API_Documentation.ipynb} (last access 29/04/2024)} are computed with the tool \emph{evo}\footnote{\url{https://github.com/MichaelGrupp/evo} (last access 29/04/2024)}, where the poses are aligned in a pre-processing step. The alignment includes rotation and translation for the robot poses and additionally scale for COLMAP poses. Note, that these metrics differ from RMST and RMSR. As MTE and MRE are influenced by the resection as well as by COLMAP / the robot, they should not be interpreted as uncertainties of the COLMAP or of the robot results only. Also, the metrics ignore the (potentially high) correlation between the elements of the poses. We have chosen to still use them in order to simplify the computations; see \cite{FoerstnerWrobel_PCV_2016}, p. 120 ff. for details about taking into account the mentioned correlations by comparing the respective covariance matrices. 

\begin{table}[t]
    \centering
    \caption{Reprojection Root Mean Squared difference (RRMS), Mean Translation Error (MTE), Mean Rotation Error (MRE), and processing time for the pose calculation with COLMAP and our approach with the robot.} 
    \begin{tabular}{lcccc}
        \textbf{Dataset} & \textbf{RRMS} $\downarrow$ & \textbf{MTE} $\downarrow$ & \textbf{MRE} $\downarrow$ & \textbf{Time} $\downarrow$\\
        & \textbf{in px} & \textbf{in mm} & \textbf{in \textdegree} & \textbf{in sec} \\
        \midrule
        GT & 0.35 & - & - & -\\ 
        \cmidrule{1-5}
        \multicolumn{5}{l}{COLMAP:}\\
        Dataset 1 & 1.13 & 0.08 & 0.051 & 10994\\
        Dataset 2 & 1.10 & 0.14 & 0.091 & 9730\\
        Dataset 3 & 1.23 & 0.08 & 0.054 & 10544\\
         Dataset 4 & 1.36 & 0.13 & 0.123 & 1897\\
        Dataset 5 & 1.93 & 3.77 & 179.037 & 1601\\
        Dataset 6 & 1.59 & 0.11 & 0.074 & 5207\\
        \cmidrule{1-5}
        Robot & 0.96 & 0.10 & 0.079 & 0.2\\
    \end{tabular}
    \label{tab:pose_acc}
\end{table}

The poses determined with the robot consistently have a high accuracy, and as opposed to those computed with COLMAP, they are obviously independent of the image content. The accuracy of the poses determined with COLMAP, on the other hand, varies. For all datasets except dataset 5, the poses are similarly accurate. 
For dataset 5, the accuracy of the poses computed with COLMAP is significantly lower than that of the robot. This is probably due to strong reflections on the light bulb in this dataset, which result in mirrored feature points from the background being used to determine the poses.
Lastly, Table \ref{tab:pose_acc} contains the processing times for COLMAP and our approach with the robot. COLMAP was processed with enabled GPU-acceleration on an NVIDIA RTX 2060 and takes between 26\,min and 3\,h for each of our datasets. Since no feature points are found in the homogeneous background of datasets 4 to 6, the related processing times are lower than for datasets 1 to 3. However, this processing time is almost completely eliminated in our approach with the robot although a non-optimized CPU implementation was used. The time measurement for our approach includes the following three steps that need to be performed for each image: i) Read joint angles, ii) apply forward kinematics to calculate the robot poses, and iii) apply hand--eye pose to calculate the camera poses.
 


\begin{figure}[t]
    \centering
    \includegraphics[width=0.6\columnwidth]{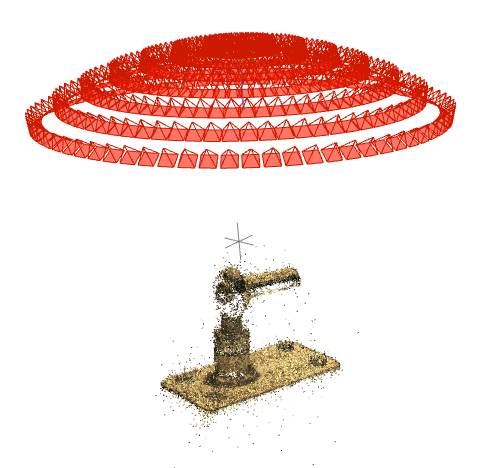}
    \vspace{-0.7cm}
    \caption{Result of COLMAP for Dataset 4, i.e.,  the reconstructed sparse point cloud and the estimated poses.}
    \label{fig:colmapResult}
\end{figure}

Figure~\ref{fig:colmapResult} visualizes the result of COLMAP for dataset 4 in the form of the poses in red and the reconstructed, sparse point cloud. The result is plausible, however, no points can be reconstructed for the homogeneous background. The poses of the images are therefore calculated based on a less favorable distribution of the image points, which explains the lower pose accuracy.

\subsection{Evaluation of the NeRF approaches}

In this subsection, we report experiments investigating the quality of the generated novel views by using different NeRF approaches from the literature and different sets of camera poses. The criteria used for the evaluation are the established metrics PSNR and SSIM, and their respective standard deviations across the evaluation images. 

From the large number of NeRF approaches that nowadays exist, we select 
Nerfacto \citep{tancik2023nerfstudio} and 3DGS \citep{kerbl20233d} for our experiments. Nerfacto is the standard method of Nerfstudio \citep{tancik2023nerfstudio} and is a representative of the methods that can be trained in a short time. We evaluate four different versions of Nerfacto: Nerfacto with and without pose refinement as explained in \cite{lin2021barf} as well as Nerfacto-big with and without pose refinement. There are four main differences between Nerfacto and Nerfacto-big: (a) The hashmap of Nerfacto-big has double the resolution (4096 vs. 2048) and four times the capacity (2M vs 500k) compared to Nerfacto; (b) Nerfacto-big trains 8192 rays per batch, in contrast to 4096 for Nerfacto; (c) Nerfacto-big uses a larger network, i.e. Nerfacto-big uses twice as many neurons per hidden layer as Nerfacto (128 vs. 64); (d) Nerfacto-big computes 100,000 iterations, whereas Nerfacto only iterates 30,000 times.

3DGS \citep{kerbl20233d} is currently considered the state-of-the-art method in terms of training and inference time as well as of visual quality of the rendered images. We use the 3DGS implementation Splatfacto from Nerfstudio \citep{tancik2023nerfstudio} to ensure a uniform evaluation. 

The various NeRF approaches are evaluated twice with different camera poses: 
\begin{enumerate}[nosep]
    \item COLMAP version: COLMAP is used to calculate the interior orientation and the camera poses in a pre-processing step that needs to be performed for each dataset separately. The COLMAP version corresponds to the conventional procedure in the NeRF community.
    \item Robot version: Like mentioned in Section \ref{sec:setup}, the camera and the hand--eye calibration are performed once in advance. The robot poses are queried from the robot's kinematics and together with the result of the hand--eye calibration, the camera poses are computed based on Equation (\ref{eq:HE_loop}). The robot version does not require a separate pre-processing step for each dataset. 
\end{enumerate} 

For 3DGS, an initial point cloud needs to be provided. For the COLMAP version, the calculation of this point cloud is straight-forward. For the robot version, we use COLMAP to calculate the sparse point cloud but fix the interior orientation and the camera poses of the robot version during optimization.

Unless otherwise specified, we use the default parameter settings of Nerfstudio for our experiments. In this way, we can evaluate the methods from the point of view of how easily they can be transferred to our industrial datasets. However, this means that higher visual quality may be achievable by tuning the parameters.

\subsection{Quantitative results of the NeRF methods}\label{sec:Results}

Table \ref{tab:NeRF_Results} shows the results of Nerfacto and 3DGS based on PSNR and SSIM and the respective standard deviations across the evaluation images, which are not part of the set of training images. The top part of the table shows the results of the COLMAP version and the bottom part the results of the robot version.
The results in Table \ref{tab:NeRF_Results} are described in more detail below under different aspects.

\begin{table*}[ht!]
    \centering
    \caption{Results and standard deviations of different NeRF methods for the COLMAP version (top) and the robot version (bottom). 'pr' means pose refinement.}
    \begin{adjustbox}{max width=\textwidth}
        \begin{tabular}{l|cc|cc|cc|cc|cc|cc}
            & \multicolumn{2}{c}{Dataset 1} & \multicolumn{2}{c}{Dataset 2} & \multicolumn{2}{c}{Dataset 3} & \multicolumn{2}{c}{Dataset 4} & \multicolumn{2}{c}{Dataset 5} & \multicolumn{2}{c}{Dataset 6} \\
            \textbf{Method} & \textbf{PSNR} $\uparrow$ & \textbf{SSIM} $\uparrow$ & \textbf{PSNR} $\uparrow$ & \textbf{SSIM} $\uparrow$ & \textbf{PSNR} $\uparrow$ & \textbf{SSIM} $\uparrow$ & \textbf{PSNR} $\uparrow$ & \textbf{SSIM} $\uparrow$ & \textbf{PSNR} $\uparrow$ & \textbf{SSIM} $\uparrow$ & \textbf{PSNR} $\uparrow$ & \textbf{SSIM} $\uparrow$ \\
            \midrule
            COLMAP version: &&&&&&&&&&&&\\
            Nerfacto w/o pr & 31.3$\pm$1.4 & 0.892$\pm$0.006 & 31.0$\pm$1.5 & 0.884$\pm$0.005 & 32.1$\pm$1.8 & 0.894$\pm$0.007 & 29.0$\pm$2.1 & 0.941$\pm$0.006 & 26.1$\pm$2.5 & 0.909$\pm$0.032 & 37.8$\pm$1.5 & 0.963$\pm$0.004 \\
            Nerfacto w/ pr   & 25.0$\pm$2.5 & 0.785$\pm$0.035 & 23.8$\pm$2.5 & 0.784$\pm$0.032 & 23.1$\pm$3.1 & 0.734$\pm$0.036 & 23.4$\pm$0.7 & 0.911$\pm$0.005 & 27.3$\pm$2.2 & 0.917$\pm$0.024 & 31.1$\pm$0.5 & 0.908$\pm$0.007 \\
            Nerfacto-big w/o pr & 30.1$\pm$1.8 & 0.909$\pm$0.005 & 29.1$\pm$2.4 & 0.895$\pm$0.010 & 32.2$\pm$2.2 &0.918$\pm$0.006 & 25.6$\pm$2.1 & 0.937$\pm$0.012 & 21.3$\pm$1.2 & 0.817$\pm$0.032 & 21.6$\pm$1.0 & 0.846$\pm$0.016\\
            Nerfacto-big w/ pr &23.9$\pm$0.9 & 0.776$\pm$0.030 & 25.8$\pm$1.3 & 0.799$\pm$0.017 & 20.3$\pm$0.9 & 0.695$\pm$0.032 & 20.2$\pm$2.2 & 0.901$\pm$0.012 & 17.6$\pm$1.0 & 0.716$\pm$0.036 & 24.2$\pm$0.9 & 0.849$\pm$0.015 \\
            3DGS & 32.1$\pm$1.5 & 0.930$\pm$0.004 & 31.8$\pm$1.5 & 0.925$\pm$0.003 & 35.1$\pm$1.2 & 0.936$\pm$0.004 & 30.4$\pm$2.2 & 0.950$\pm$0.005 & 27.0$\pm$1.4 & 0.925$\pm$0.007 & 41.0$\pm$1.1 & 0.978$\pm$0.002\\
            \midrule
            Robot version: &&&&&&&&&&&&\\
            Nerfacto w/o pr & 31.3$\pm$1.4 & 0.883$\pm$0.009 & 30.1$\pm$1.4 & 0.870$\pm$0.012 & 31.3$\pm$1.8 & 0.876$\pm$0.012 & 26.6$\pm$2.1 & 0.933$\pm$0.007 & 33.3$\pm$1.5 & 0.951$\pm$0.006 & 37.1$\pm$1.0 & 0.957$\pm$0.004 \\
            Nerfacto w/ pr   & 25.1$\pm$2.5 & 0.783$\pm$0.042 & 25.2$\pm$1.5 & 0.791$\pm$0.028 & 23.2$\pm$3.3 & 0.734$\pm$0.050 & 22.8$\pm$1.1 & 0.904$\pm$0.005 & 28.1$\pm$2.6 &  0.938$\pm$0.011 & 31.1$\pm$1.5 & 0.909$\pm$0.013 \\
            Nerfacto-big w/o pr & 29.9$\pm$1.5 & 0.894$\pm$0.008 & 28.9$\pm$2.2 & 0.881$\pm$0.016 & 31.1$\pm$2.1 & 0.893$\pm$0.016 & 24.9$\pm$2.2 & 0.927$\pm$0.012 & 24.8$\pm$1.9 & 0.886$\pm$0.033 & 21.2$\pm$0.7 & 0.823$\pm$0.014 \\
            Nerfacto-big w/ pr & 22.5$\pm$2.8  & 0.756$\pm$0.039 & 26.5$\pm$2.3 & 0.812$\pm$0.022 & 23.5$\pm$2.0 & 0.731$\pm$0.033 & 23.3$\pm$2.4 & 0.914$\pm$0.014 & 23.4$\pm$1.6 & 0.866$\pm$0.024 & 25.2$\pm$1.2 & 0.882$\pm$0.022 \\
            3DGS & 32.0$\pm$1.3 & 0.923$\pm$0.006 & 31.5$\pm$1.6 & 0.915$\pm$0.015 & 34.3$\pm$1.3 & 0.923$\pm$0.013 & 30.1$\pm$2.0 & 0.958$\pm$0.005 & 33.3$\pm$1.8 & 0.965$\pm$0.004 & 40.3$\pm$1.0 & 0.975$\pm$0.003 \\
        \end{tabular}
    \end{adjustbox}
    \label{tab:NeRF_Results}
\end{table*}

\textbf{COLMAP vs. Robot.} The results of the COLMAP and the robot version do not differ significantly with one exception: For dataset 5, PSNR and SSIM are considerably higher for the robot versions for all methods. The reason is the higher accuracy of camera poses of the robot version compared to COLMAP (see Table \ref{tab:pose_acc}). This highlights the disadvantage of using SfM to determine the camera poses in cases where the image content is challenging for SfM. Using the robot for pose estimation is more robust and provides consistent accuracy, whereas with COLMAP the accuracy strongly depends on the image content. 

\textbf{Best performing methods.} On all datasets, 3DGS achieves higher values for PSNR and SSIM than all Nerfacto methods while having faster training and inference times. Nerfacto w/o pose refinement achieves the second-best results. Interestingly, Nerfacto-big achieves worse results than Nerfacto, even though it uses a larger network and trains for more iterations. Apparently, the size of Nerfacto’s network is already sufficient to represent the given complexity of the industrial scenes. With the exception of dataset 5, the choice of the method has a larger impact on the evaluation metrics than the way in which the poses are determined.

\textbf{Pose refinement.} The comparison of the Nerfacto methods with and without pose refinement clearly shows that the pose refinement leads to a considerable deterioration for our industrial datasets. The poses, which can already be determined very accurately with both COLMAP and the robot, become less accurate due to the simultaneous optimization during NeRF training, which leads to lower PSNR and SSIM values. When looking at the rendered images, difference between the results with and without pose refinement are small. Nevertheless, it can be seen that the variant with pose refinement renders images that are rotated and shifted in relation to the reference image. We therefore argue that the pose refinement is responsible for the degradation. We attribute this finding to the higher ambiguity during training of the methods with simultaneous pose refinement and the already high accuracy of the input poses.

\subsection{Qualitative results of the NeRF methods}
In general, the visual quality of the rendered images reflects the results in Table \ref{tab:NeRF_Results} very well. For this reason, a small number of images are selected here for the purpose of illustration.

\begin{figure}[t]
  \centering
  \begin{subfigure}{0.49\columnwidth}
    \includegraphics[width=\textwidth]{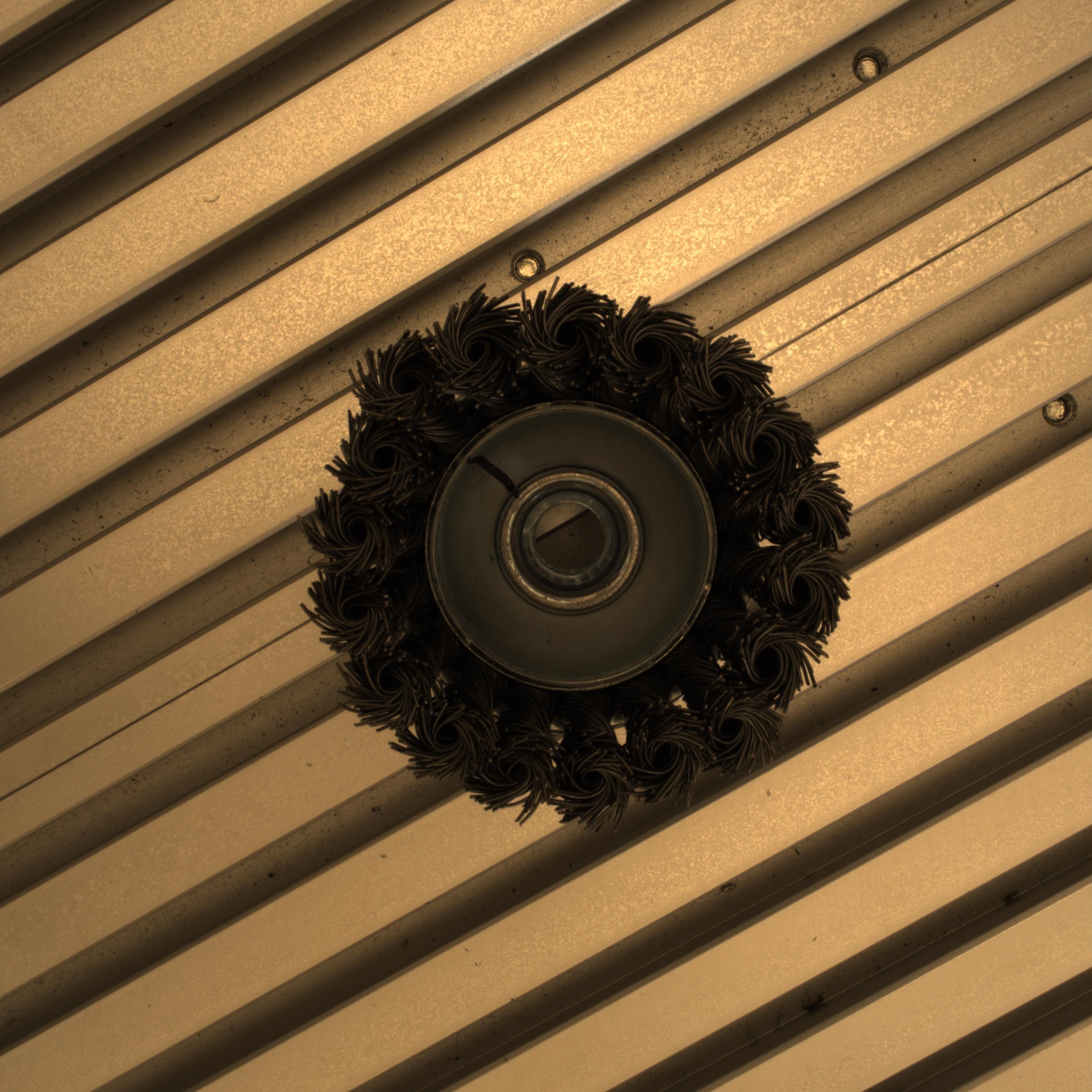}
    \caption{Reference view}
  \end{subfigure}
  \hfill
  \begin{subfigure}{0.49\columnwidth}
    \includegraphics[width=\textwidth]{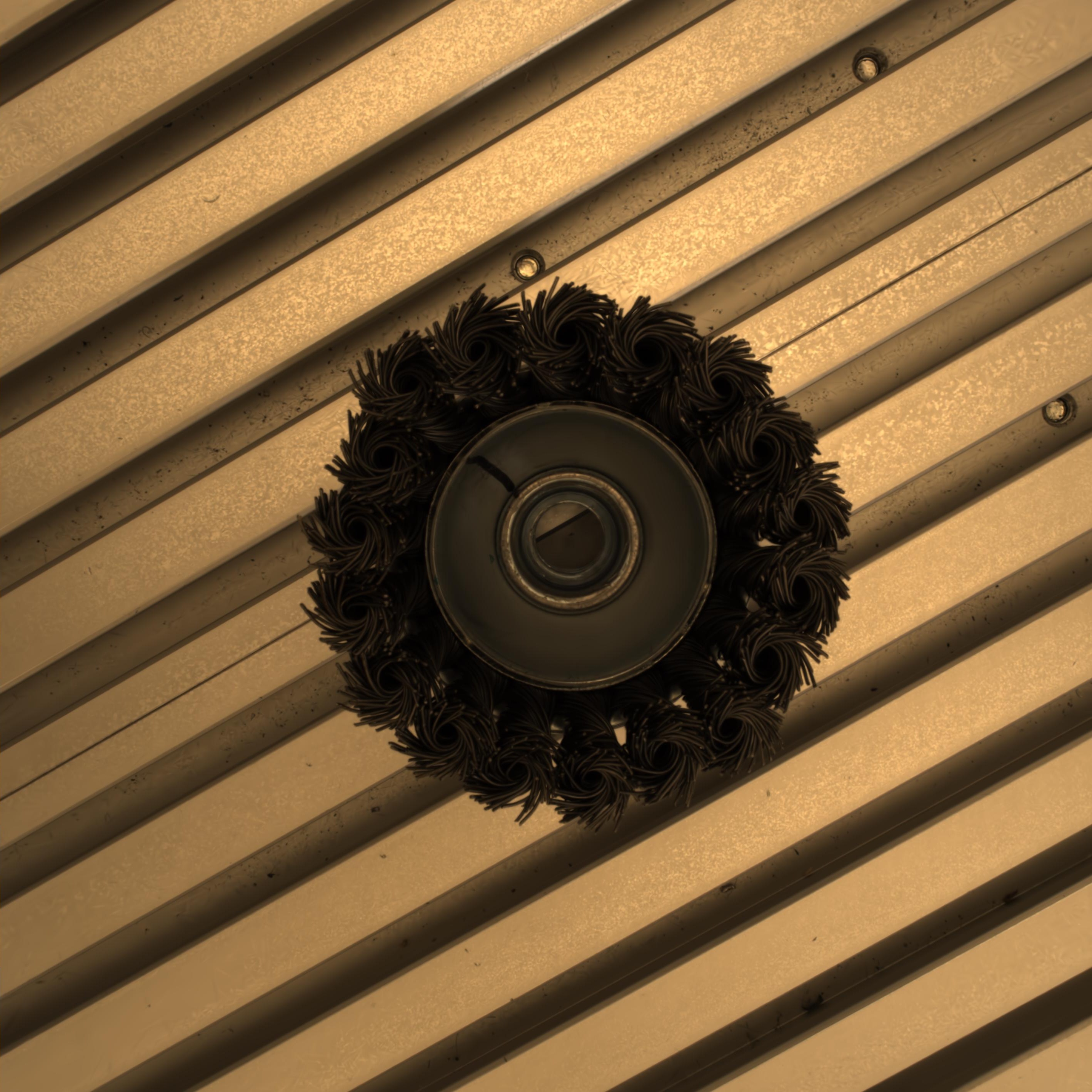}
    \caption{3DGS}
  \end{subfigure}
  \caption{Reference view and rendered result of the robot version of 3DGS for an image of dataset 3.}
  \label{fig:Splatfacto_D3_robot}
\end{figure}

Figure~\ref{fig:Splatfacto_D3_robot} visualizes the reference view and the rendered image of the robot version of 3DGS for an evaluation image of dataset 3. With 3DGS, extraordinarily realistic images can be rendered for this dataset. There is no recognizable difference to the reference view. Consequently, the NeRF methods are capable of densifying an industrial dataset with realistically rendered images.

\begin{figure}[t]
  \centering
  \begin{subfigure}[t]{0.49\columnwidth}
    \includegraphics[width=\textwidth,angle=180,origin=c]{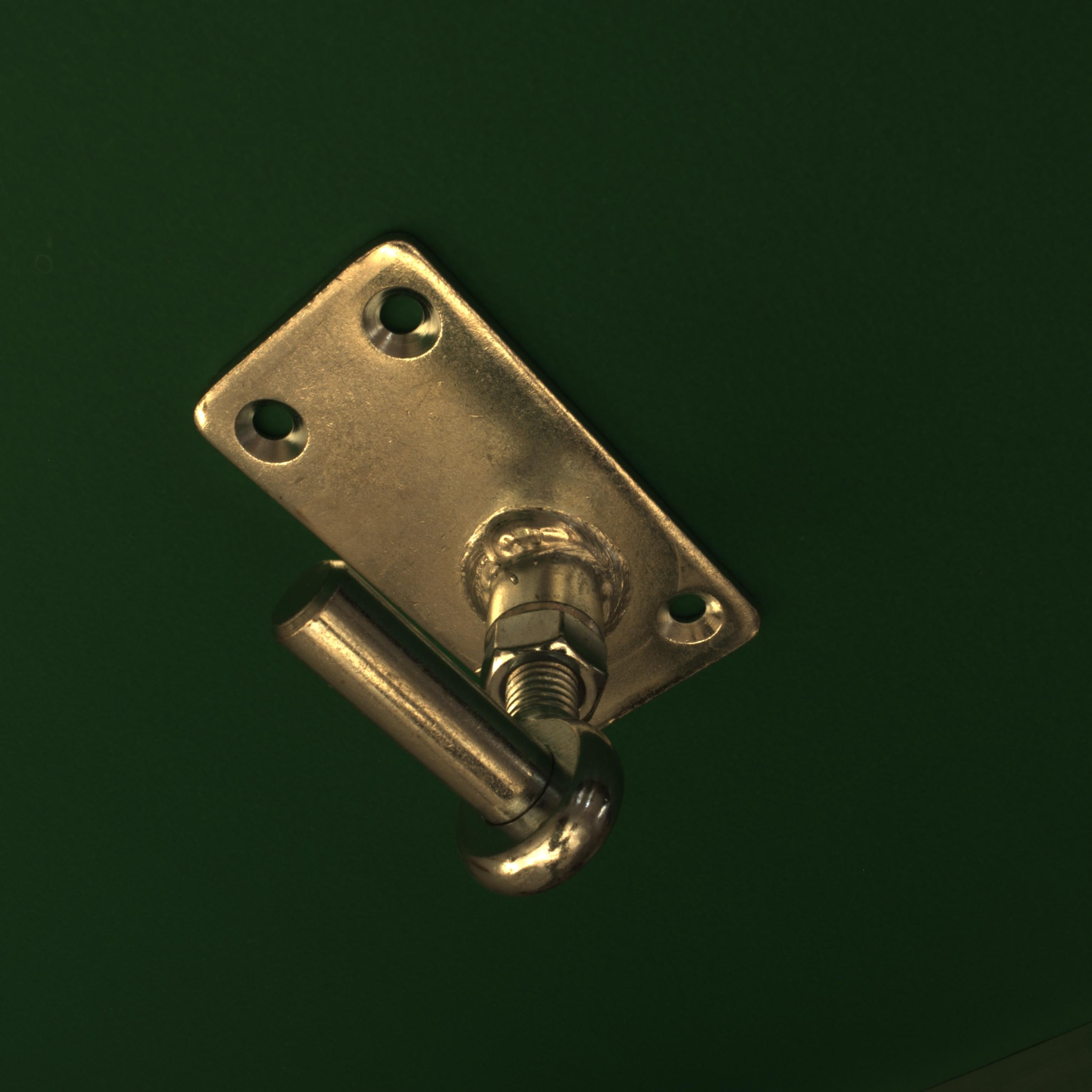}
    \caption{Reference view}
  \end{subfigure}
  \hfill  
  \begin{subfigure}[t]{0.49\columnwidth}
    \includegraphics[width=\textwidth,angle=180,origin=c]{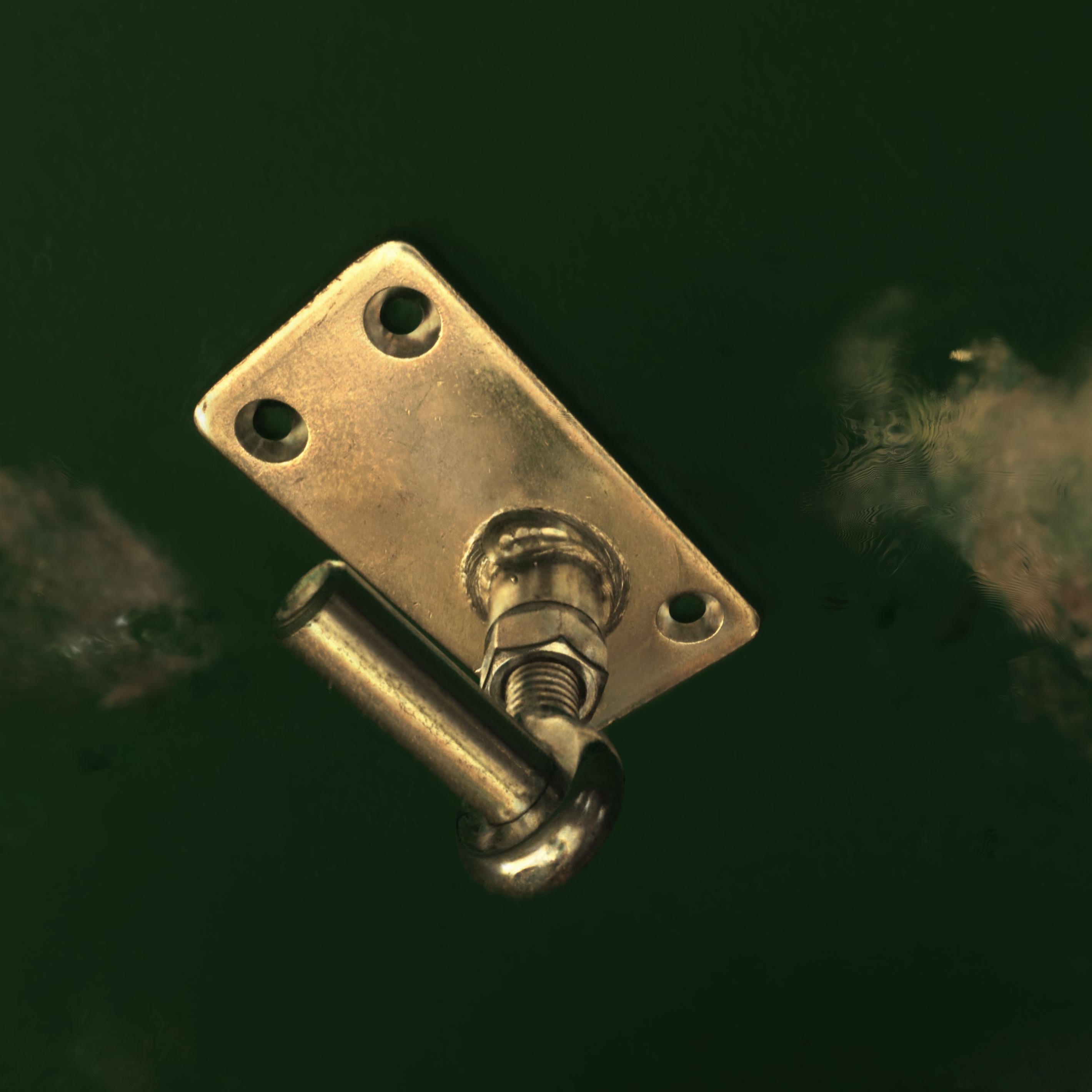}
    \caption{Nerfacto-big w/o pose refinement}
  \end{subfigure}
  \caption{Reference view and rendered result of the COLMAP version of Nerfacto-big without pose refinement for an image of dataset 4.}
  \label{fig:Poserefinement}
\end{figure}

Figure~\ref{fig:Poserefinement} shows the reference view and the rendered image of the COLMAP version of Nerfacto-big without pose refinement for an image of dataset 4. In this case, the image rendered by Nerfacto-big contains artifacts. In addition, the reflections, which are generally a challenge, are not reproduced completely correctly.

\begin{figure}[t]
  \centering
  \begin{subfigure}[t]{0.49\columnwidth}
    \includegraphics[width=\textwidth,angle=180,origin=c]{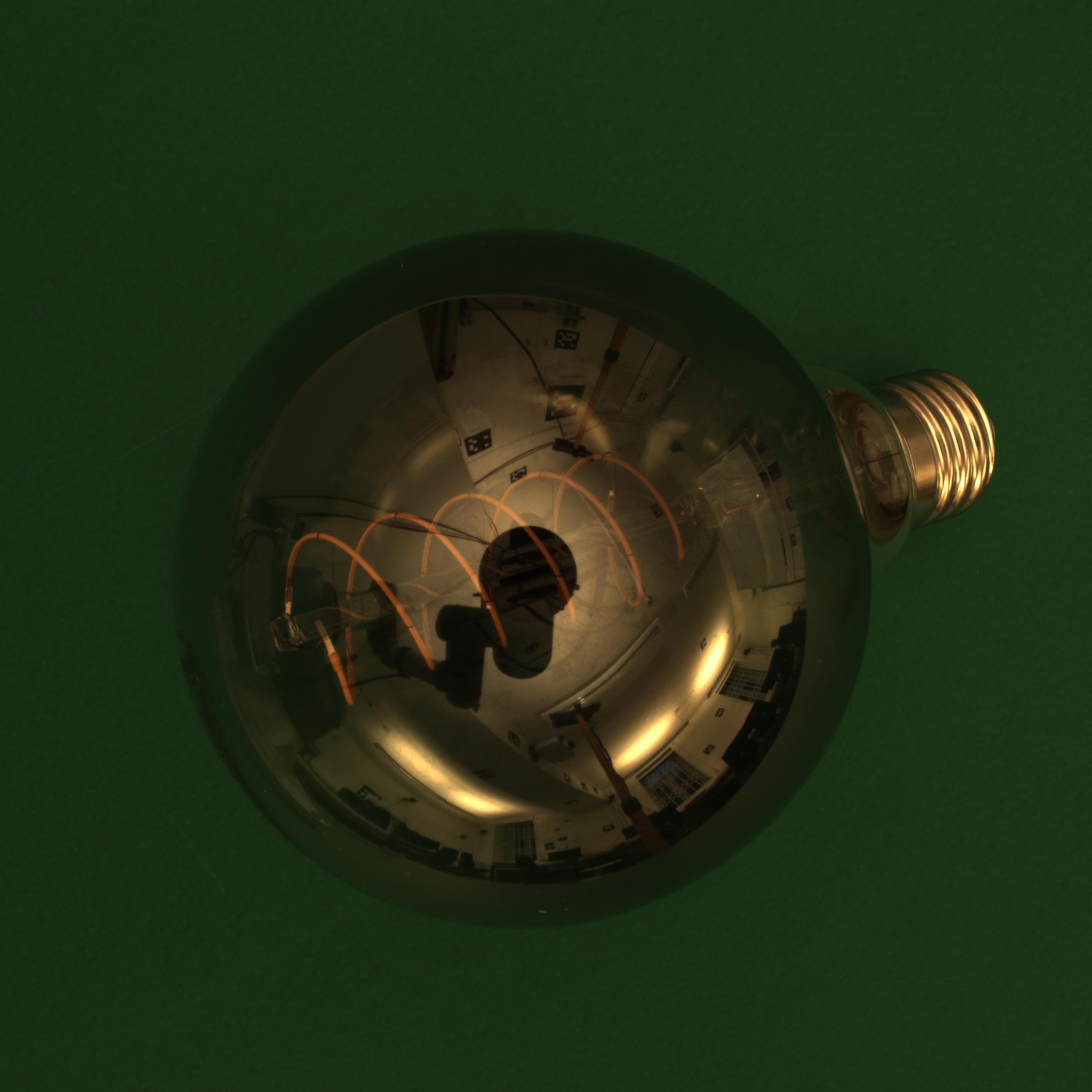}
    \caption{Reference view}
  \end{subfigure}

  \begin{subfigure}[t]{0.49\columnwidth}
    \includegraphics[width=\textwidth,angle=180,origin=c]{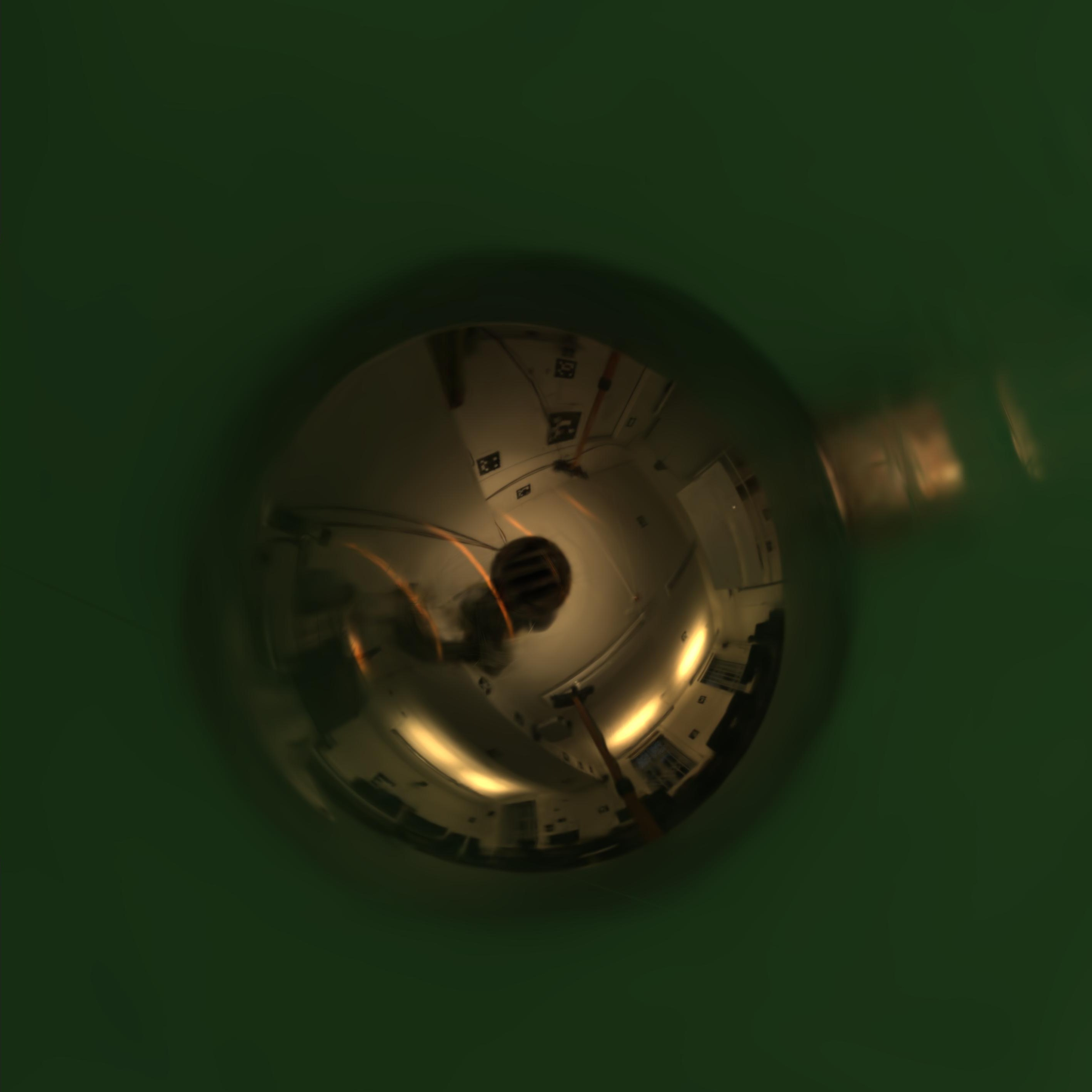}
    \caption{COLMAP version of 3DGS}
  \end{subfigure}
  \hfill
  \begin{subfigure}[t]{0.49\columnwidth}
    \includegraphics[width=\textwidth,angle=180,origin=c]{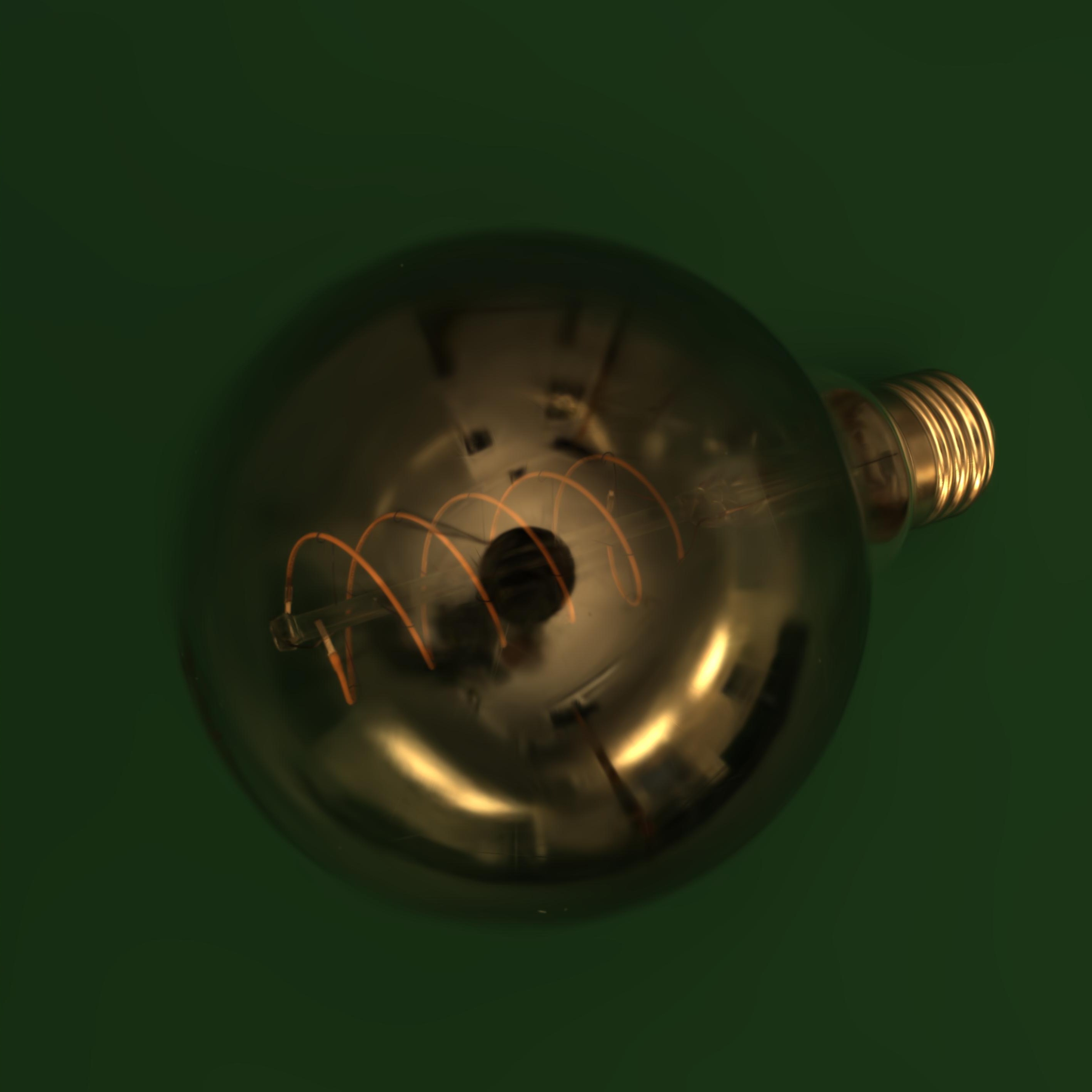}
    \caption{Robot version of 3DGS}
  \end{subfigure}  
  \caption{Reference view and rendered results of the COLMAP and the robot version of 3DGS for an image of dataset 5.}
  \label{fig:Dataset5}
\end{figure}

Figure~\ref{fig:Dataset5} visualizes the reference view as well as the results of the COLMAP and the robot version for an image of dataset 5. Exemplary results from 3DGS are shown, but the same applies to the other methods. The COLMAP version of 3DGS and also of the other methods struggle to model the base of the bulb. Interestingly, at the same time, the reflections of the background are reconstructed in fine detail. We assume that COLMAP has found a plausible solution for calculating poses for the reflected image areas. These poses then lead to a good reconstruction of the reflected background in the NeRFs. For the area on the object that is not reflected, however, the poses do not represent a plausible solution. 
In contrast, for the robot version of 3DGS, the base of the bulb can be reconstructed and the rendered image looks very similar to the reference view, which matches the higher PSNR and SSIM values in Table \ref{tab:NeRF_Results}. However, the reflections of the background look slightly blurrier. Although the results for this difficult object are impressive, this example shows that strong reflections are still a major challenge for novel view synthesis, even for the state-of-the-art method.

\subsection{Uncertainty quantification with a NeRF ensemble}
In order to investigate the uncertainty of novel view generation using NeRF approaches, in the following experiments, the simplified NeRF ensemble approach as described in \citep{sunderhauf2023} is applied exemplarily to dataset 1. Each of the $M$ ensemble members is a Nerfacto \citep{tancik2023nerfstudio} model, with pose optimization disabled and otherwise default hyperparameters. Randomness is achieved through different initialization of the learnable parameters. 

Then, the ensemble prediction of the color $\hat{\mathbf{c}}_E(\mathbf{r})$, $\mathbf{c} = (r, g, b)$ of a ray $\mathbf{r}$ is the arithmetic mean across all ensemble members' color predictions $\hat{\mathbf{c}}_m(\mathbf{r})$ for this ray.
\begin{equation}
    \hat{\mathbf{c}}_E(\mathbf{r}) = \frac{1}{M} \sum^M_{m=1} \hat{\mathbf{c}}_m(\mathbf{r})
\end{equation}
Similarly, the uncertainty is expressed as the standard deviation of the individual predictions, aggregating the three color channels:
\begin{equation}
    \sigma_{\mathbf{c}_E}(\mathbf{r}) = \sqrt{\frac{1}{3M} \sum_{m=1}^M \Delta \mathbf{c}_{m,E}^2(\mathbf{r})}
\end{equation}
with the residuals between ensemble prediction and reference view
\begin{equation}
    \Delta \mathbf{c}_{m,E}^2(\mathbf{r}) = \sum_{i=1}^3 \left(\hat{\mathbf{c}}_{m,i}(\mathbf{r}) - \hat{\mathbf{c}}_{E}(\mathbf{r})\right)^2.
\end{equation}

\subsubsection{In-distribution experiment.}
For evaluation of the uncertainty quantification, each ensemble member ($M=30$) is trained using a subset of dataset 1 with 100 regularly distributed images, but otherwise independently of each other. The evaluation images are located in-between training views, i.e. in-distribution.

\begin{figure}[t]
  \centering
  \begin{subfigure}[b]{0.49\columnwidth}
    \includegraphics[width=\textwidth,angle=180,origin=c]{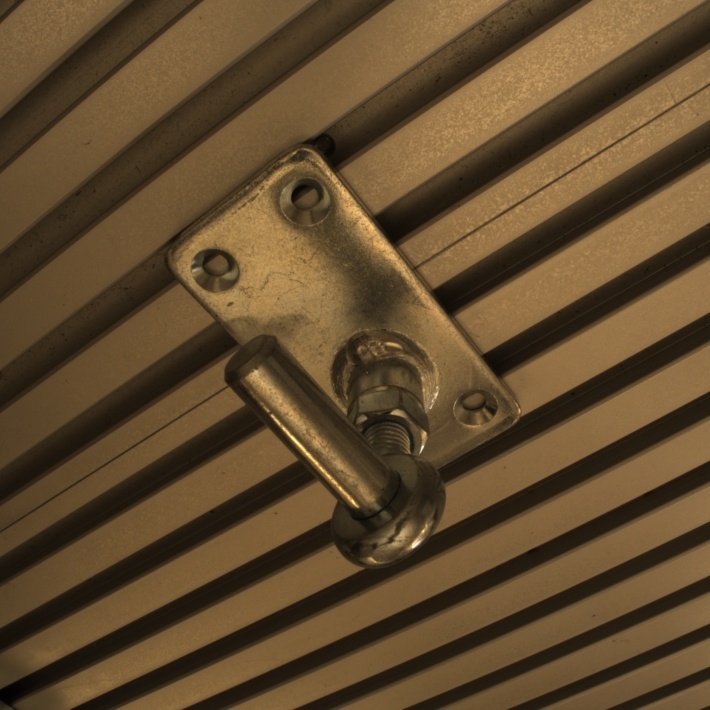}
    \caption{Reference view}
  \end{subfigure}
  \hfill
  \begin{subfigure}[b]{0.49\columnwidth}
    \includegraphics[width=\textwidth,angle=180,origin=c]{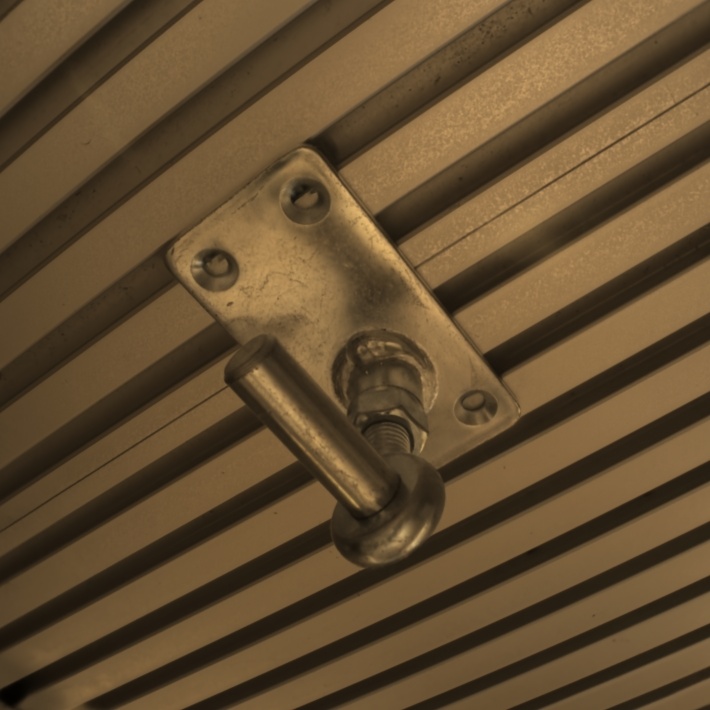}
    \caption{Ensemble prediction}
  \end{subfigure}
  
  \begin{subfigure}[b]{0.49\columnwidth}
    \includegraphics[width=\textwidth,angle=180,origin=c]{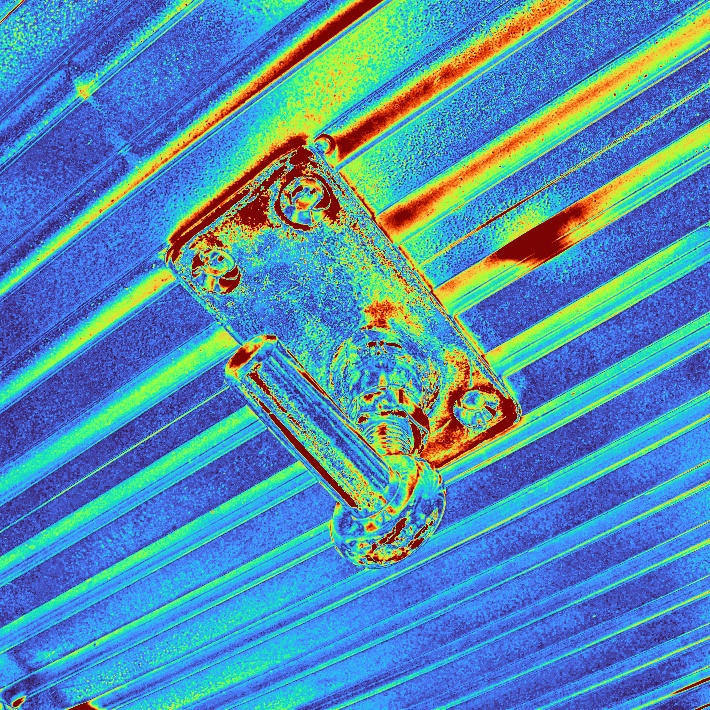}
    \caption{Magnitude of residuals}
   \end{subfigure}
   \hfill
   \begin{subfigure}[b]{0.49\columnwidth}
    \includegraphics[width=\textwidth,angle=180,origin=c]{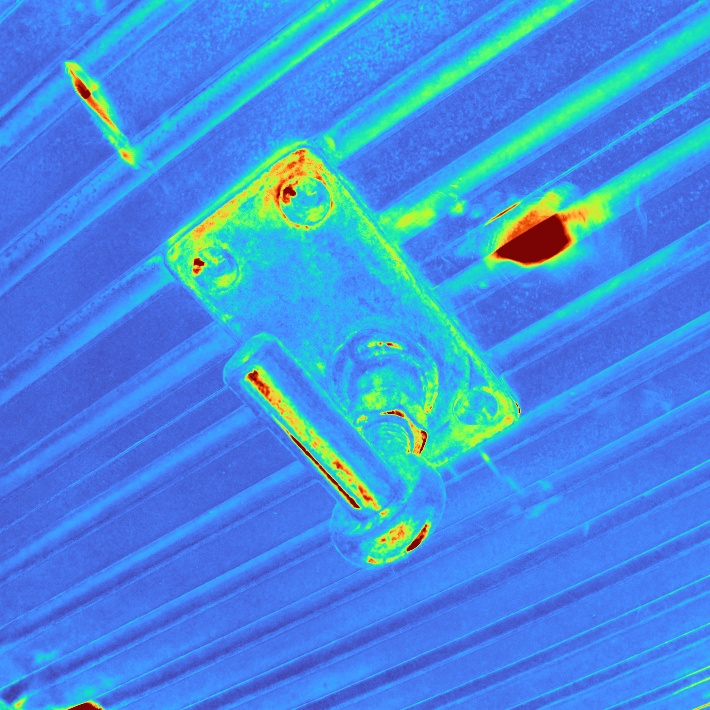}
    \caption{Ensemble standard deviation}
  \end{subfigure}

  \begin{subfigure}[b]{0.49\columnwidth}
    \includegraphics[width=\textwidth,angle=180,origin=c]{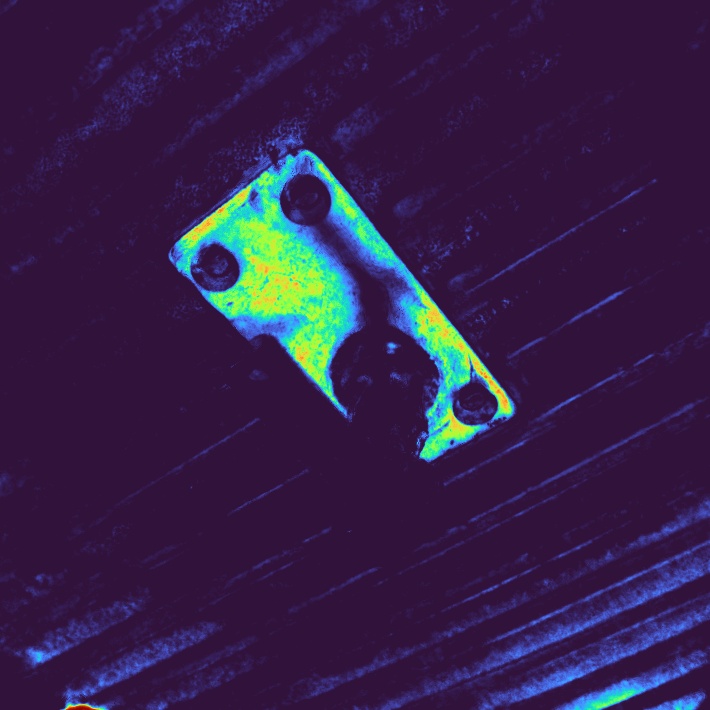}
    \caption{Inverse accumulated density}
  \end{subfigure}

  \caption{NeRF ensemble results for an image of dataset 1.}
  \label{fig:exp1uncertainty}
\end{figure}

Figure~\ref{fig:exp1uncertainty} shows an example ensemble prediction of an evaluation image, allowing for a qualitative evaluation. Although the shape appears to be captured well, the color prediction (b) exhibits fewer details than the reference view (a), resulting in a smoothed appearance. This is reflected in the residuals $\Delta\mathbf{c}_{E,ref}$ (c) appearing grainy. The cause can be found in the individual members' predictions, which also appear smoothed, indicating that the default Nerfacto model lacks capacity. Comparing (c) and (d), the residuals visually correlate with the predicted standard deviation to some degree. 
For example, the oblong specular reflection on the edge of the cylinder visible in (a) is less pronounced in the prediction, leading to large residuals, but also a large standard deviation where members are in disagreement. A second area of high standard deviation can be found to the left of the object. This is caused by the presence of a floater in a single member's prediction, which can be considered an outlier. This violates the assumption of normally distributed colors.

It is apparent that the predicted standard deviation is too low in many areas of large residuals. Prominently, the edge of the base of the object and the grooves of the aluminum profile in the background exhibit very large residuals, but only moderate standard deviations. The ensemble is over-confident. \citet{sunderhauf2023} propose to extend the uncertainty equation by the squared inverse accumulated density (visualized in Figure~\ref{fig:exp1uncertainty}(e), different color scale) to account for unobserved structure. In our case, the comparatively lower density of the base plate does not coincide well with the large residuals.

\subsubsection{Out-of-distribution experiment.}
In this experiment, only training views located in one half of the dome are used in order to evaluate the ensemble's performance with out-of-distribution samples. With a total of 60 views for ensemble training, $M=10$ ensemble members are trained with slightly different subsets of 50 training views each.

\begin{figure}[t]
  \centering
  \begin{subfigure}[b]{0.49\columnwidth}
    \includegraphics[width=\textwidth,angle=180,origin=c]{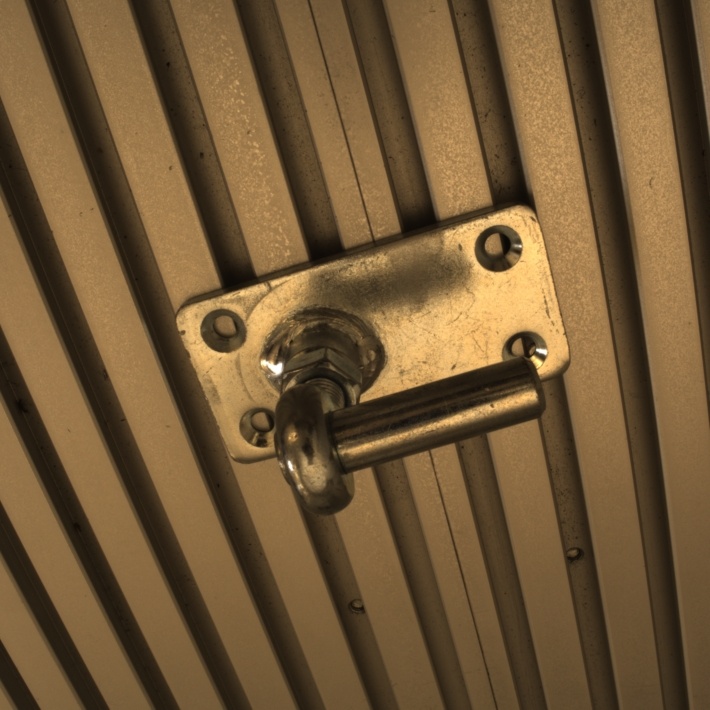}
    \caption{Reference view}
  \end{subfigure}
  \hfill
  \begin{subfigure}[b]{0.49\columnwidth}
    \includegraphics[width=\textwidth,angle=180,origin=c]{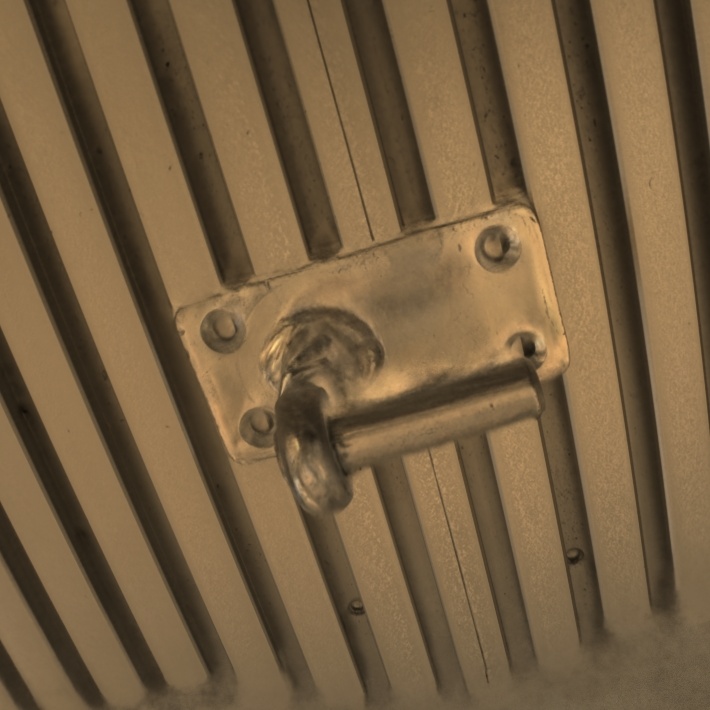}
    \caption{Ensemble prediction}
  \end{subfigure}
  
  \begin{subfigure}[b]{0.49\columnwidth}
    \includegraphics[width=\textwidth,angle=180,origin=c]{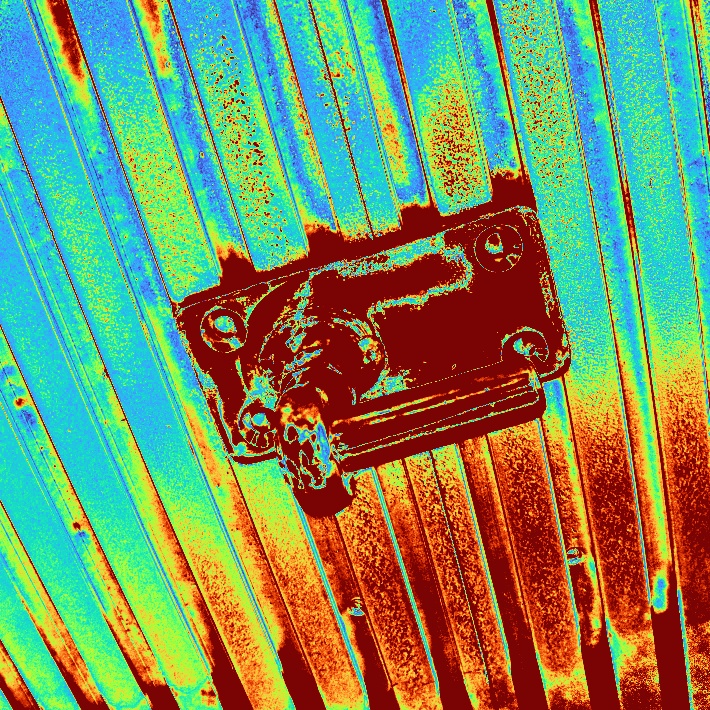}
    \caption{Magnitude of residuals}
   \end{subfigure}
   \hfill
   \begin{subfigure}[b]{0.49\columnwidth}
    \includegraphics[width=\textwidth,angle=180,origin=c]{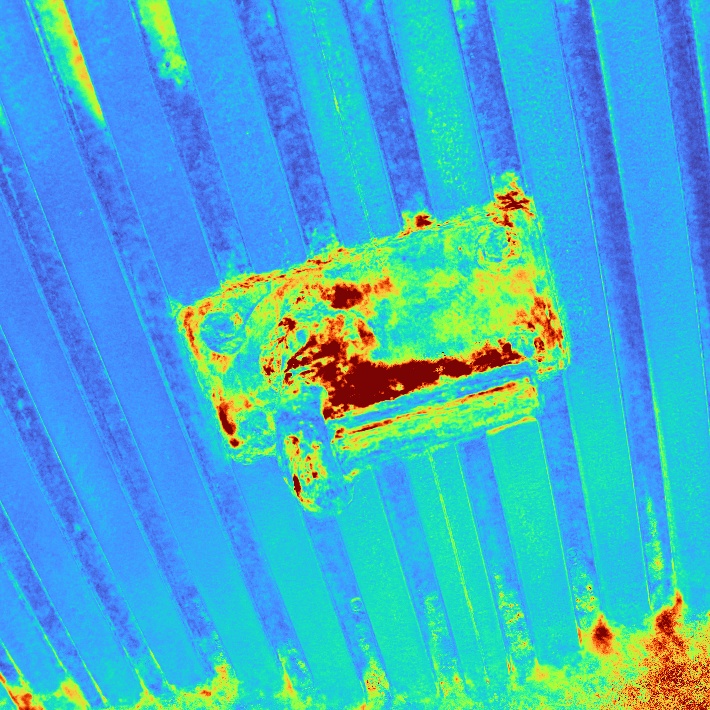}
    \caption{Ensemble standard deviation}
  \end{subfigure}

  \caption{Ensemble prediction on out-of-distribution evaluation image}
  \label{fig:exp2uncertainty}
\end{figure}

Figure~\ref{fig:exp2uncertainty} shows the ensemble prediction for an example evaluation image that is located opposite the training views, i.e. on the back side. Remarkably, the prediction (b) nevertheless shows that the general shape and color of the scene have been captured to some degree. However, parts of the scene that were unobserved during training exhibit artifacts, and the surface lacks significantly in detail.
View-dependent effects like specular reflection can be observed to cause large residuals. These effects are not captured in the standard deviation: The strong view-dependent component of the radiance field at the reflective surface was underfit to a similar degree in each ensemble member. This is a limitation of modeling the distribution in image space.
On the other hand, the patch of large predicted standard deviation stems from inconsistent 3D structure, a typical artifact in unobserved regions of a NeRF and thus a desired outcome.

\section{Conclusion}\label{sec:Conclusion}

In this paper, we evaluated the potential of novel view synthesis with NeRFs for industrial robot applications. We captured the input images with a camera attached to the end effector of an industrial robot and determined accurate camera poses based on the robot kinematics. Our datasets contain typical challenges for industrial applications, like reflecting objects, poor texture, and fine structures, and allow us to explore the limits of the conventional approach of using SfM to determine the camera poses.
Our experiments show that with careful calibration and an accurate robot, this approach achieves comparable quality to the conventional method. However, our approach is more robust and leads to consistently high quality of images rendered by the NeRFs, whereas the conventional approach requires image content favorable for image-based determination of the camera poses. Furthermore, our approach determines the camera poses with metric scale, which favors potential subsequent steps in which metric information is required. Lastly, it is faster when flexible adjustments of the poses are required, which is a critical factor across various applications.

Since many industrial applications require a high degree of reliability, we further quantify uncertainties of the rendered images. In principle, a simplified ensemble approach provides the correct tendencies for the uncertainties for in-distribution views, as shown by the correlation with the residuals. However, the ensemble’s predicted standard deviation is often too low in areas of large residuals, indicating overconfidence. Furthermore, the experiment on out-of-distribution poses shows that NeRFs are, in principle, able to generate novel views from perspectives unseen during training. However, these views exhibit fewer details in texture and structure, and view-dependent effects like specular reflection cause large residuals. These effects are not captured in the ensembles' standard deviation, which demonstrates the limitations of the ensemble approach.

In the future, we will extend our analysis from novel view synthesis to the task of 3D scene or object reconstruction, respectively, in the industrial context. Furthermore, we will also look into the thermal domain for industrial applications. Since determining the camera poses and interior orientation with SfM is expected to be more challenging for thermal images, we consider the potential of using a robot for pose estimation to be even higher when applied to e.g. multi-modal NeRF approaches \citep{poggi2022xnerf} in order to enable cross-spectral scene representation and analysis.


{
	\begin{spacing}{1.17}
		\normalsize
		\bibliography{main} 
	\end{spacing}
}

\end{document}